\documentclass[nohyperref]{article}

\usepackage{microtype}
\usepackage{graphicx}
\usepackage{subfigure}
\usepackage{booktabs} 
\usepackage{multirow}
\usepackage{hyperref}

\usepackage[accepted]{icml2022}

\usepackage{amsmath}
\usepackage{amssymb}
\usepackage{mathtools}
\usepackage{amsthm}
\usepackage{bm}
\usepackage{bbm}
\usepackage[capitalize,noabbrev]{cleveref}

\theoremstyle{plain}

\theoremstyle{definition}

\theoremstyle{remark}

\usepackage[textsize=tiny]{todonotes}

\definecolor{darkred}{rgb}{0.7,0,0}
\definecolor{darkgreen}{rgb}{0,0.46,0}
\definecolor{purple}{rgb}{0.6,0,0.5}

\newcommand{\tens}[1]{\mathcal{#1}}

\newcommand{\matr}[1]{\boldsymbol{#1}}

\newcommand{\vect}[1]{\boldsymbol{#1}}

\newcommand{\RR}{\mathbb{R}}

\newcommand{\W}{\mathcal{W}}
\newcommand{\M}{\mathcal{M}}

\icmltitlerunning{Data-free Backdoor Removal based on Channel Lipschitzness}

\begin{document}

\twocolumn[
\icmltitle{Data-free Backdoor Removal based on Channel Lipschitzness}



\icmlsetsymbol{equal}{*}

\begin{icmlauthorlist}
\icmlauthor{Runkai Zheng*}{cuhksds}
\icmlauthor{Rongjun Tang*}{cuhksse}
\icmlauthor{Jianze Li, Li Liu}{sribd}
\end{icmlauthorlist}

\icmlaffiliation{cuhksds}{School of Data Science, The Chinese University of Hong Kong (Shenzhen), Shenzhen, China}
\icmlaffiliation{cuhksse}{School of Science and Engineering, The Chinese University of Hong Kong (Shenzhen), Shenzhen, China}
\icmlaffiliation{sribd}{Shenzhen Research Institute of Big Data, Shenzhen, China}

\icmlcorrespondingauthor{Li Liu}{liuli@cuhk.edu.cn}

\icmlkeywords{Machine Learning, ICML}

\vskip 0.3in
]



\printAffiliationsAndNotice{\icmlEqualContribution} 

\begin{abstract}
Recent studies have shown that Deep Neural Networks (DNNs) are vulnerable to the backdoor attacks, which leads to malicious behaviors of DNNs when specific triggers are attached to the input images. It was further demonstrated that the infected DNNs possess a collection of channels, which are more sensitive to the backdoor triggers compared with normal channels. Pruning these channels was then shown to be effective in mitigating the backdoor behaviors. To locate those channels, it is natural to consider their Lipschitzness, which measures their sensitivity against worst-case perturbations on the inputs. In this work, we introduce a novel concept called Channel Lipschitz Constant (CLC), which is defined as the Lipschitz constant of the mapping from the input images to the output of each channel. Then we provide empirical evidences to show the strong correlation between an Upper bound of the CLC (UCLC) and the trigger-activated change on the channel activation. Since UCLC can be directly calculated from the weight matrices, we can detect the potential backdoor channels in a data-free manner, and do simple pruning on the infected DNN to repair the model. The proposed Channel Lipschitzness based Pruning (CLP) method is super fast, simple, data-free and robust to the choice of the pruning threshold. Extensive experiments are conducted to evaluate the efficiency and effectiveness of CLP, which achieves state-of-the-art results among the mainstream defense methods even without any data. Source codes are available at https://github.com/rkteddy/channel-Lipschitzness-based-pruning.

\end{abstract}

\section{Introduction}

In recent years, Deep Neural Networks (DNNs) have achieved significantly advanced performance in a wide range of fields, and have been further applied to industrial practices, including some security-critical fields, \textit{e.g.}, audio-visual processing \cite{liu2020re} and medical image processing \cite{chen2021uscl}. Accordingly, the model security problems have gained much attention from the community.

One of the well-known model security problems of DNN is the adversarial attack \cite{tramer2018ensemble}, \cite{madry2018towards}, \cite{biggio2011support}, which has been extensively studied. The main idea is to add imperceptible perturbation into a well-classified sample to mislead the model prediction during the testing phase. Recently, backdoor attacks \cite{gu2019badnets} remarkably show severe threats to model security due to its well-designed attack mechanism, especially under high safety-required settings. Different from adversarial attacks, backdoor attacks modify the models by a carefully designed strategy during the model training process. For example, the attackers may inject a small proportion of malicious training samples with trigger patterns before training. In this way, DNN can be manipulated to have designated responses to inputs with specific triggers, while acting normally on benign samples \cite{gu2019badnets}.

Backdoor attacks may happen in different stages of the model adopting pipeline \cite{li2020backdoor}. 
For example, when the victims try to train a DNN model on a suspicious third-party dataset, the implanted poisoned backdoor data perform the backdoor attack silently. The defender under such scenario thus has full access to the training process and the whole training data. Besides, backdoor attack happens with a greater frequency when the training process is uncontrollable by the victim, such as adopting third-party models or training on a third-party platform. Under such setting, the defender is only given a pre-trained infected model and usually a small set of benign data as an additional auxiliary to help to mitigate the backdoored model.

\begin{figure*}[t]
    \centering
    \includegraphics[width=0.6\linewidth]{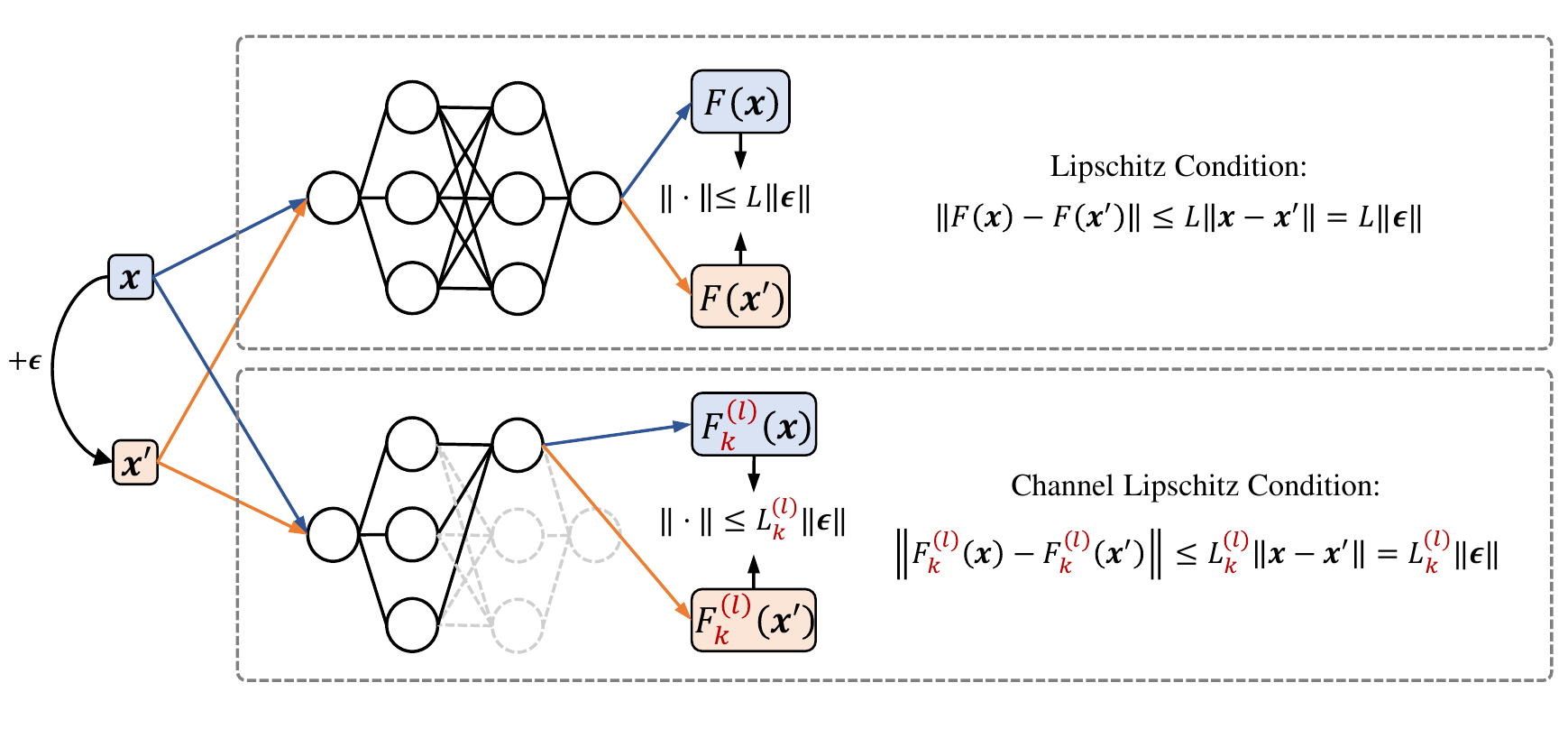}
    \caption{An illustration of the differences between the commonly studied Lipschitz condition and the proposed channel Lipschitz condition. We highlight the differences in the formula with red color. In a nutshell, Lipschitz constant bounds the largest changing rate of the whole network function, while channel Lipschitz constant bounds that of a specific output channel.}
    \label{fig:clc}
\end{figure*}

To better understand the mechanism of backdoor attacks, in this work, we revisit the common characteristics of backdoor attacks, \textit{i.e.}, the ability of a small trigger to flip the outputs of benign inputs into the malicious target label. It is natural to relate such sensitivity to the high Lipschitz constant of the model. However, it may be a too strict condition for controlling the Lipschitz constant of the whole network function to achieve backdoor robustness. Since the backdoor vulnerability may come from the sensitivity of the channels to the backdoor trigger, we instead consider evaluating the Lipschitz constant (strictly speaking, an upper bound of the Lipschitz constant) among the channels to identify and prune the sensitive channels.

Specifically, we view the mapping from the input images to each channel output as an independent function, as shown in \cref{fig:clc}. The Lipschitz constant is considered for each channel, which measures the sensitivity of the channel to the input. As channels, which detect backdoor triggers, should be sensitive to specific perturbation (\textit{i.e.}, the trigger) on inputs, 
\textbf{we argue that backdoor related channels should have a high Lipschitz constant compared to normal channels}.
To demonstrate this, we track the forward propagation process of the same inputs with and without trigger to see how the trigger changes the activation on these channels. We provide empirical evidence to show the strong correlation between the Channel Lipschitz Constant (CLC) and the trigger-activated changes. To be more specific, we show that channels with large activation changes after attaching the trigger usually have high Lipschitz constant. Intuitively, pruning those channels may mitigate the changes brought by the backdoor trigger. 

To this end, we propose a novel \emph{Channel Lipschitzness based Pruning} (CLP), which prunes the channels with high Lipschitz constant to recover the model. Since the CLC can be directly derived from the weight matrices of the model, this method requires no access to any training data.
Unlike previous methods that are designed for different specific threat models, \textbf{CLP's data-free property ensures a high degree of flexibility for its practical application, as long as the model parameters are accessible}. Besides, CLP is super fast and robust to the choice of its only hyperparameter, \textit{i.e.}, the relative pruning threshold $u$.
Finally, we show that CLP can effectively reduce the attack success rate (ASR) against different advanced attacks with only a negligible drop on the clean accuracy.

To summarize, our contributions are twofold:

1. We innovatively reveal the connections between backdoor behaviors and the channels with high Lipschitz constants. This conclusion generalizes various backdoor attacks, shedding light on the development of backdoor defense.

2. Inspired by the above observations, we propose the CLP, a data-free backdoor removal method. Even without data, it achieves state-of-the-art (SOTA) performance among existing defense methods which require a certain amount of benign training data. 

\section{Related work}

\subsection{Backdoor Attack} 
BadNets \cite{gu2019badnets} is one of the earliest researches that perform backdoor attacks in DNNs. They injected several kinds of image pattern, referred to as the \emph{triggers}, into some samples, and modified the labels of the these samples to the desired malicious label. 
Then the DNNs were trained with the poisoned dataset and will be implanted a backdoor. After that, more covert and advanced backdoor designs were proposed \cite{turner2019label,xue2020one,liu2020reflection}. 
To prevent defenders from reversely generating possible triggers through the model, dynamic backdoor attacks such as the Input-Aware Backdoor (IAB) \cite{nguyen2020input}, Warping-based Backdoor (WaNet) \cite{nguyen2021wanet} and Sample Specific Backdoor Attack (SSBA) \cite{li2021invisible} were proposed. They generate a unique trigger for every single input, which makes the defense even more difficult. 
These attacks can be classified as the poisoning based backdoor attacks.

Under some settings, it is also possible for the attackers to directly modify the architectures and parameters of a DNN, without poisoning any training data, known as the \emph{non-poisoning based backdoor attacks}.
For example, in Trojan attack \cite{Trojan}, the triggers were optimized to increase the activation of some carefully chosen neurons, and then the network was retrained to finish the poisoning. 
In target bit trojan \cite{rakin2020tbt} and ProFlip \cite{chen2021proflip}, the authors choose vulnerable neurons and apply bit-flip techniques to perform backdoor attacks. 
Note that, such bit flip attack only occurs in the deployment phase of DNNs, and the ASR does not exceed other traditional ones. Therefore, this attack will not be discussed in this paper. 

\subsection{Backdoor Defense}

\subsubsection{Training Stage Defenses.}
The \emph{training stage defenses} aim at suppressing the effect of implanted backdoor triggers or filtering out the poisoned data during the training stage, with a full access to the training process. 
According to the different distributions of poisoned data and clean data in feature space \cite{huang2021backdoor}, several methods were proposed to filter out the poisoned data, including the robust statistics \cite{SPECTRE,tran2018spectral}, input perturbation techniques \cite{gao2019strip,doan2020februus} and semi-supervise training \cite{huang2021backdoor}. 
Besides, stronger data augmentation techniques \cite{borgnia2021strong} were proposed to suppress the effect of backdoor, such as the CutMix \cite{devries2017improved}, CutOut \cite{devries2017improved} and MaxUp \cite{gong2020maxup}. Besides, differential privacy \cite{du2019robust} and randomized smoothing \cite{rosenfeld2020certified} provide some certificated methods to defend against backdoor attack. 



\subsubsection{Model Post-processing Defenses.}
The \emph{model post-processing defenses} mainly focus on eliminating the backdoor threat in a suspicious DNN model. 
The first effective defense against backdoor attack with the combination of neuron pruning and fine-tuning was proposed in  \cite{liu2018fine}.  
Inspired by the mechanism of fixed-trigger backdoor attack, Neural Cleanse (NC) \cite{wang2019neural} obtained the reversed engineering trigger and detoxify the model with such knowledge. Some other fine-tuning based defenses used knowledge distillation \cite{hinton2015distilling} in their pipeline, such as the backdoor knowledge distillation \cite{yoshida2020disabling} and Neural Attention Distillation (NAD) \cite{li2020neural}. Besides, the Mode Connectivity Repair \cite{zhao2019bridging} was also explored to eliminate the backdoor effect. 

In addition to fine-tuning based defenses, the $L_{\inf}$\footnote{Refers to the infinite norm.} Neuron Pruning was proposed in \cite{xu2020defending} to filter out the neurons with high $L_{\inf}$ in activations. 
Recently, the Adversarial Neuron Pruning (ANP) \cite{wu2021adversarial} detected and pruned sensitive neurons with adversarial perturbations, and achieve considerable defense performance. However, it needs careful tuning of hyperparameters and requires access to a certain number of benign data.

Unlike those model post-processing methods, the proposed CLP achieves superior defending performance without any benign data, and is robust to the choice of the \textbf{only} hyperparameter. Moreover, we hope that our work can enlighten the study on the effectiveness of Lipschitz constant on backdoor learning, and provide a new perspective on backdoor attack and defense.
\section{Preliminaries}

\subsection{Notations}

In this paper, we consider a classification problem with $C$ classes. 
Suppose that $\mathcal{D}=\{(\vect{x}_i, y_i)\}_{i=1}^{N}\subseteq\mathcal{X}\times\mathcal{Y}$ is the original training set, which contains $N$ i.i.d. sample images $x_i \in \mathbb{R}^{d_c \times d_h \times d_w}$ and the corresponding labels $y_i \in \{1,2,...,C\}$. 
Here, we denote by $d_c$, $d_h$ and $d_w$ the number of output and input channels, the height and the width of the image, respectively. 
It is clear that $d_c=3$ for RGB images.


We consider a neural network $F(x; \theta)$ with $L$ layers:
\begin{equation}\nonumber
F(x; \theta)=f^{(L)} \circ \phi \circ f^{(L-1)} \circ \cdots \circ \phi\circ  f^{(1)},
\end{equation}
where $f^{(l)}$ is the linear function (\emph{e.g.,} convolution) in the $l^{th}$ layer with $1\leq l\leq L$, $\phi$ is a non-linear activation function applied element wise. For simplicity, we denote $F(x; \theta)$ as $F(x)$ or $F$. 
Let $\tens{W}^{(l)}\in \mathbb{R}^{d_{c'}^{(l)}\times d_c^{(l)} \times d_h^{(l)} \times d_w^{(l)}}$
be the weight tensor of the $l^{th}$ convolutional layer, where $d_{c'}^{(l)}, d_{c}^{(l)}, d_{h}^{(l)}$ and $d_{w}^{(l)}$ are the number of output and input channels, the height and the width of the convolutional kernel, respectively. To do pruning, we apply a mask $\tens{M}^{(l)}\in \{0, 1\}^{d_{c'}^{(l)}\times d_c^{(l)} \times d_h^{(l)} \times d_w^{(l)}}$
starting with $\tens{M}^{(l)}_k=\mathbf{1}_{d_{c}^{(l)} \times d_h^{(l)} \times d_w^{(l)}}$
in each layer, where $\mathbf{1}_{d_c^{(l)} \times d_h^{(l)} \times d_w^{(l)}}$ 
denotes an all-one tensor. Pruning neurons on the network refers to getting a collection of indices $\mathcal{I}=\{(l, k)_i\}_{i=1}^{I}$ and setting $\M^{(l)}_k=\mathbf{0}_{d_c^{(l)} \times d_h^{(l)} \times d_w^{(l)}}$ if $(l,k)\in \mathcal{I}$. The pruned network $F_{-\mathcal{I}}$ has the same architecture as $F$ with all the weight matrices of convolutional layers set to $\W^{(l)}\odot \M^{(l)}$, where $\odot$ denotes the Hadamard product \cite{horn2012matrix}.

The backdoor poisoning attack involves changing the input images and the corresponding labels\footnote{the labels remain unchanged in clean label attacks \cite{turner2019label}.} on a subset of the original training set $\mathcal{D}_p\subseteq \mathcal{D}$. 
We denote the poisoning function to the input images by $\delta(x)$. The ratio $\rho = \frac{\vert \mathcal{D}_p \vert}{\vert \mathcal{D} \vert}$ is defined as the \emph{poisoning rate}.

\subsection{$L$-Lipschitz Function}
A function $g:\RR^{n_1}\xrightarrow[]{}\RR^{n_2}$ is \emph{$L$-Lipschitz continuous} \cite{armijo1966minimization} in $\mathcal{X}\subseteq\mathbb{R}^{n_1}$ under $p$-norm, if there exists a non-negative constant $L\geq0$ such that 
\begin{equation} \label{eq:lips_g}
\Vert g(\vect{x})-g(\vect{x}') \Vert_p \leq L\|\vect{x}-\vect{x}'\|_p, \quad \forall\ \vect{x}, \vect{x}' \in \mathcal{X}.
\end{equation}
The smallest $L$ guaranteeing equation \eqref{eq:lips_g} is called the \emph{Lipschitz constant} of $g$, denoted by $\Vert g \Vert_{\rm Lip}$. 
For simplicity, we choose $p=2$ in this paper. 
Viewing $\vect{x}'$ as a perturbation of $\vect{x}$, the Lipschitz constant $\Vert g \Vert_{\rm Lip}$ can be regarded as the maximum ratio between the resulting perturbation in the output space and the source perturbation in the input space. Thus, it is commonly used for measuring the sensitivity of a function to the input perturbation.

\begin{figure}[t]
    \centering
    \includegraphics[width=\linewidth]{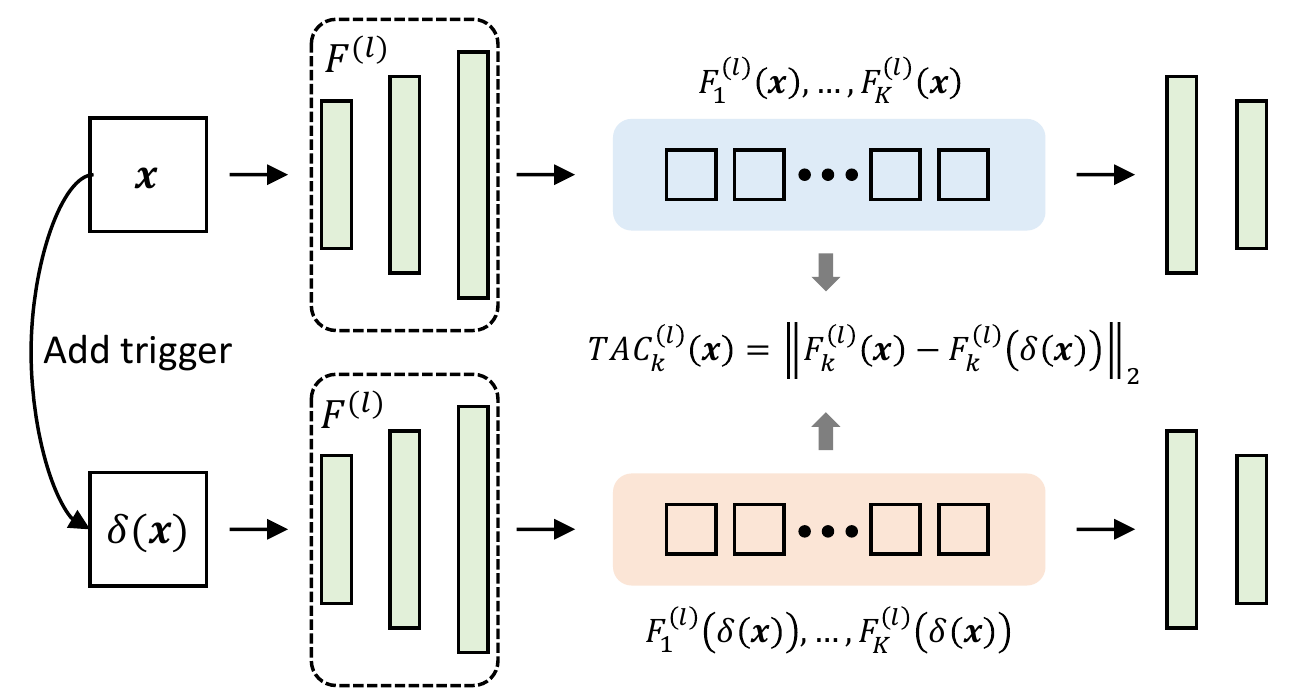}
    \caption{A simple diagram for calculating TAC in the $l^{th}$ layer. As illustrated, the word TAC stands for the activation differences of the feature maps before and after attaching the trigger to the images, and $k$ is the number of feature maps.}
    \label{fig:tac}
\end{figure}

\subsection{Lipschitz Constant in Neural Networks}

According to the Cauchy-Schwartz inequality, we are now able to control the Lipschitz constant of the whole network as follows:
\begin{align}
\Vert F \Vert_{\rm Lip} &= \Vert f^{(L)} \circ \phi \circ f^{(L-1)} \circ \cdots \circ \phi\circ  f^{(1)} \Vert_{\rm Lip} \nonumber\\
&\leq \Vert f^{(L)} \Vert_{\rm Lip} \cdot \Vert \phi\Vert_{\rm Lip} \cdot \Vert f^{(L-1)} \Vert_{\rm Lip}  \cdots \Vert \phi\Vert_{\rm Lip}\cdot \Vert f^{(1)} \Vert_{\rm Lip}. 
\end{align}
Most of the commonly used activation functions are L-Lipschitz (\emph{e.g.,} ReLU, LeakyReLU, Sigmoid, Tanh, ELU, SeLU). 
In particular, we have $L=1$ for the ReLU function, which is used in this paper. 
Note that $f^{(l)}(\vect{x}^{(l)})=\matr{W}^{(l)} \vect{x}^{(l)}+\vect{b}^{(l)}$.
It follows that 
\begin{align}
\Vert F \Vert_{\rm Lip} \leq \prod_{l=1}^{L} \Vert f^{(l)} \Vert_{\rm Lip}
= \prod_{l=1}^{L} \max_{\Vert \vect{z} \Vert_{2} \neq 0} \frac{\Vert \matr{W}^{(l)}\vect{z} \Vert_{2}}{\Vert \vect{z} \Vert_{2}} = \prod_{l=1}^{L} \sigma(\matr{W}^{(l)}),
\end{align}
where $\sigma(\matr{W}^{(l)}) = \max_{\Vert \vect{z} \Vert_{2} \neq 0} \frac{\Vert \matr{W}^{(l)}\vect{z} \Vert_{2}}{\Vert \vect{z} \Vert_{2}}$ is the spectral norm.

\section{Methodology}

\begin{figure}[t]
    \centering
    \includegraphics[width=\linewidth]{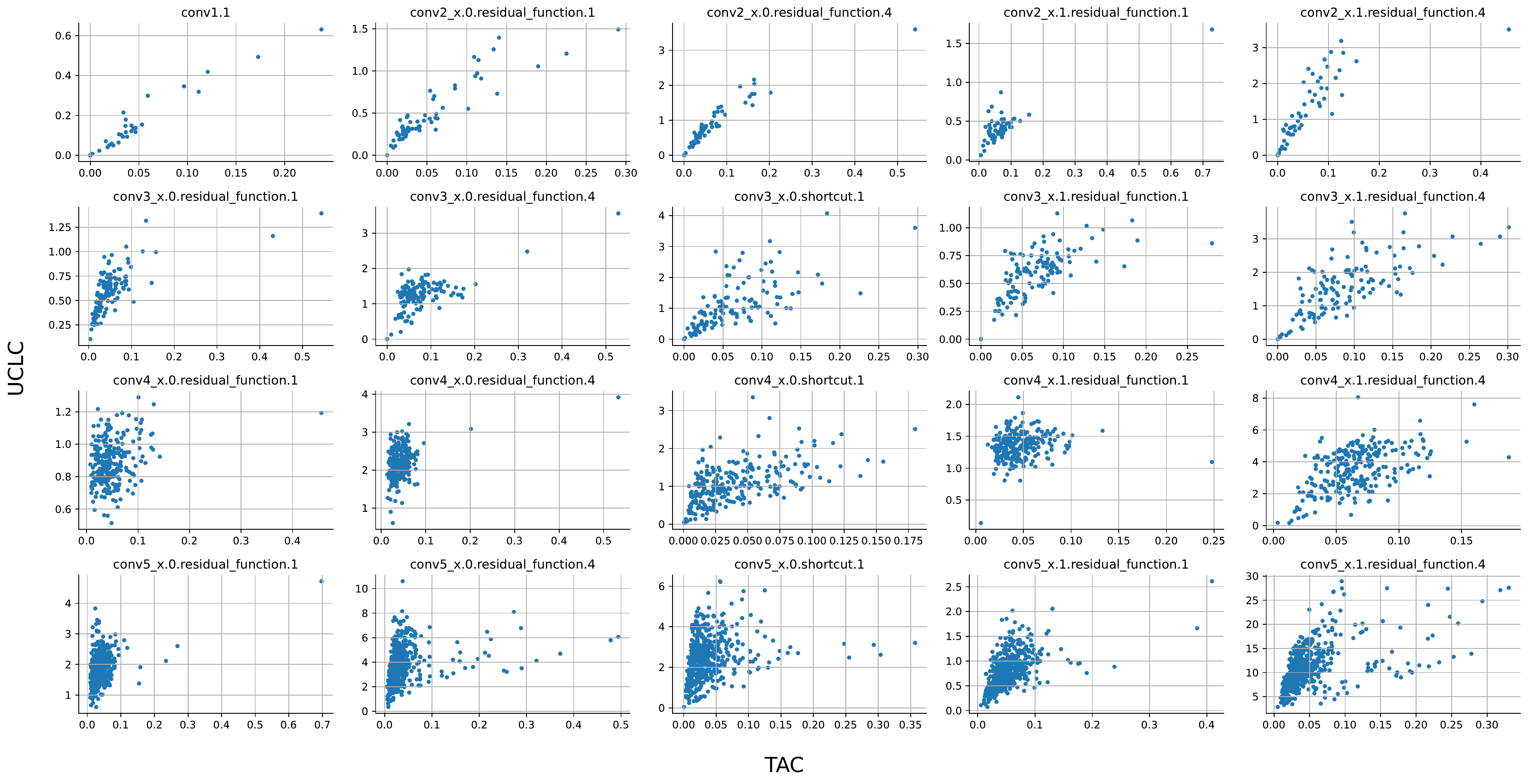}
    \caption{A scatter plot to demonstrate the relationship between UCLC and TAC. As shown in the figure, we observe a strong correlation between the two indices.}
    \label{fig:clc_tac}
\end{figure}

\subsection{Channel Lipschitz Constant}
We denote the function from the input to the $k^{th}$ channel of $l^{th}$ layer by:
\begin{equation}
    F^{(l)}_k = f_{k}^{(l)} \circ \phi \circ f^{(l-1)} \cdots \circ \phi \circ f^{(1)}, \nonumber
\end{equation}
where $f^{(l)}_k$ is the $k^{th}$ output channel function of the $l^{th}$ layer. In particular, if 
\begin{equation}
    f^{(l)}(\vect{x}) = \matr{W}^{(l)}\vect{x} + \vect{b}^{(l)}, \quad \matr{W}^{(l)}\in \mathbb{R}^{d_{\rm out}^{(l)}\times d_{\rm in}^{(l)}}, \nonumber
\end{equation}
then $f^{(l)}_{k}(\vect{x}) = \vect{w}^{(l)}_k\vect{x} + b^{(l)}_{k}$, 
where $\vect{w}^{(l)}_k\in \mathbb{R}^{1 \times d_{in}^{(l)}}$ is the $k^{th}$ row of $\matr{W}^{(l)}$. It follows that the \emph{channel Lipschitz constant} (CLC) satisfies 
\begin{equation}
    \Vert F^{(l)}_k \Vert_{\rm Lip} \leq \|\vect{w}^{(l)}_k\|_{\rm Lip} \prod_{i=1}^{l-1} \sigma(\matr{W}^{(i)}). \nonumber
\end{equation}
We refer to the right side of the above inequality as an \emph{Upper bound} of the Channel Lipschitz Constant (UCLC).

Particularly, convolution is a special linear transformation, with the weight matrix $\matr{W}^{(l)}$ being a doubly block-Toeplitz (DBT) form of the weight tensor $\tens{W}^{(l)}$. To calculate the exact spectral norm of a convolution operation, one should first convert the weight tensor into a DBT weight matrix.
However, calculating the DBT matrix is time-consuming and memory-expensive, especially when the number of channels is large. 
A much simpler alternative way for this is to reshape the weight tensor into a matrix form and use the spectral norm of the reshaped matrix as an approximation to the original spectral norm. This approximation has been adopted by previous researches \cite{yoshida2017spectral}. 
In our work, for simplicity, we calculate the spectral norm using the reshaped kernel matrix $W^{(l)}_k\in\mathbb{R}^{d^{(l)}_c\times (d^{(l)}_hd^{(l)}_w)}$, which also shows acceptable results in our experiments.


\subsection{Trigger-activated Change}
In order to study the extent to which these channels are related to the backdoor behaviors, we define the \emph{Trigger-activated Change} (TAC). Specifically, we first train a ResNet-18 with BadNets in CIFAR-10, and track the forward propagation process of the same inputs with and without trigger. 
TAC is defined as the average differences of the $k^{th}$ channel of the $l^{th}$ layer for test samples  $\vect{x}$ in dataset $\mathcal{D}$:
\begin{equation}
    TAC^{(l)}_k(\mathcal{D}) = \frac{1}{\vert \mathcal{D} \vert} \sum_{\vect{x}\in\mathcal{D}} \Vert F^{(l)}_k(\vect{x}) - F^{(l)}_k(\delta(\vect{x})) \Vert_2, \nonumber
\end{equation}
where $\delta(\cdot)$ is the poisoning function. 
A more detailed illustration of this quantity is shown in \cref{fig:tac}. 
TAC is the change of the activation before and after the input image is attached with a trigger. Its magnitude reflects the effect the trigger has on a given channel. 
Higher TAC indicate higher sensitivity of the channel to the trigger. Note that TAC is proposed to study the backdoor behavior, but it can not used for defensing. This is because the calculation of TAC requires the access to the trigger pattern, which defenders cannot get in general.

\subsection{Correlation between CLC and TAC}
A scatter plot chart of TAC and UCLC for each layer under BadNets (All to One) \cite{gu2019badnets} is shown in \cref{fig:clc_tac},
from which we can observe a high correlation between them. 
In particular, there are some outlying channels with extremely high TAC in some layers, indicating that they are sensitive to the trigger. Hence, it is reasonable to consider them as the potential backdoor channels. As expected, most of these channels have abnormally high UCLC. Remind that TAC is inaccessible for a defender, but UCLC can be directly calculated from the weight matrices of the given model. Hence, we use UCLC as an alternated index to detect potential backdoor channels. We will show that pruning those high-UCLC channels will significantly reduce the backdoor ASR in \cref{sec: exp}. 

\subsection{Special Case in CNN with Batch Normalization}
The \emph{Batch Normalization} (BN) \cite{ioffe2015batch} is adopted to stabilize the training process and make optimization landscape much smoother in modern neural networks. It normalizes the batch inputs of each channel and adjusts the mean and variance through trainable parameters. BN is usually placed after the convolution and before the non-linear function. Note that BN is also a linear transformation, and can be merged with convolution into a matrix-matrix multiplication. In this paper, we view a Conv-BN block as one linear layer, and an UCLC is calculated based on the composed weight matrix. 

\subsection{Channel Lipschitzness based Pruning}
\label{sec:CLP}
Based on the above observations, it is natural to think of removing the high-UCLC channels to recover the model. On this basis, we propose the channel Lipschitzness based pruning (CLP), which calculates UCLC for channels in each layer, and prunes the channels with UCLC larger than a pre-defined threshold within the layer. 
Note that in the same layer, all channels share the same cumulative product term, which is the Lipschitz upper bound of the previous layers. Hence, a much simplified way to compare CLC within a particular layer is to directly compare $\sigma(\matr{W}^{(l)}_k)$. The overall algorithm procedure is shown in Algorithm \cref{alg:clp}.

Determining an appropriate threshold is crucial to the performance of this method. In this work, we simply set the threshold for the $l^{th}$ layer as $\mu^{(l)} + u*s^{(l)}$, where $\mu^{(l)}=\frac{1}{K}\sum_{i=1}^K \sigma(\matr{W}^{(l)}_k)$ and $s^{(l)}=\sqrt{\frac{1}{K}\sum_{i=1}^K (\sigma(\matr{W}^{(l)}_k) - \mu^{(l)})^2}$ are the mean and the standard deviation of the $l^{th}$ layer indices set $\{\sigma(\matr{W}^{(l)}_k): k=1, 2, \dots K\}$, and $u$ is the only hyperparameter for the CLP. The above threshold is also known as the common outlier dividing line for a standard Gaussian distribution. We find this simple setting works well in our experiments.

\begin{algorithm}[tb]
    \caption{Channel Lipschitzness based Pruning}
    \label{alg:clp}
    \begin{algorithmic}
        \STATE {\bfseries Input:} L layer neural network function $F^{(L)}$ with a set of convolution weight tensor $\{\tens{W}^{(l)}: l=1,2,\dots,L\}$, threshold hyperparameter $u$.
        \FOR{l = 1 to L}
            \FOR{k = 1 to K}
                \STATE $\matr{W}^{(l)}_k := \operatorname{ReshapeToMatrix}(\tens{W}^{(l)}_k)$
                \STATE $\sigma^{(l)}_k := \sigma(\matr{W}^{(l)}_k)$
            \ENDFOR
            \STATE $\mu^{(l)} :=\frac{1}{K}\sum_{i=1}^K \sigma^{(l)}_k$
            \STATE $s^{(l)} :=\sqrt{\frac{1}{K}\sum_{i=1}^K (\sigma^{(l)}_k - \mu^{(l)})^2}$
            \STATE $\mathcal{I}^{(l)} := \{(l, k): \sigma^{(l)}_k > \mu^{(l)} + u*s^{(l)}\}$
            \STATE $\mathcal{I} = \mathcal{I} \cup \mathcal{I}^{(l)}$
        \ENDFOR
        \STATE {\bfseries Output:} A pruned Network $F^{(L)}_{-\mathcal{I}}$
    \end{algorithmic}
\end{algorithm}

\section{Experiments}
\label{sec: exp}
\subsection{Experimental Settings}
\subsubsection{Attack Settings.} We test our proposed CLP on a variety of famous attack methods, \textit{i.e.}, BadNets \cite{gu2019badnets}, Clean Label Attack \cite{turner2019label}, Trojan Attack \cite{Trojan}, Blended Backdoor Attack \cite{liu2020reflection}, WaNet \cite{nguyen2021wanet}, IAB \cite{nguyen2020input} and Sample Specific Backdoor Attack \cite{li2021invisible}. The attacks are performed on CIFAR-10 \cite{krizhevsky2009learning} and Tiny ImageNet \cite{le2015tiny} using ResNet-18 \cite{he2016deep}. For BadNets, we test both All-to-All (BadA2A) attack and All-to-One (BadA2O) attack, which means that the attack target labels $y_t$ are set to all labels by $y_t=(y+1) \% C$ (\% denotes modulo operation) or one particular label $y_t=C_t$. Due to the image size requirement of the SSBA, its corresponding experiments are only conducted on Tiny ImageNet. We use $\sim 95\%$ of the training data to train the backdoored model. The rest $5\%$ are split into $4\%$ of validation data, and $1\%$ of benign training data to perform other defenses. The trigger size is $3 \times 3$ for CIFAR-10 and $5 \times 5$ for Tiny ImageNet. The poison label is set to the $0^{th}$ class, and the poisoning rate is set to $10\%$ by default. We use SGD \cite{ruder2016overview} as the base optimizer, and train the backdoored model with learning rate 0.1, momentum 0.9 and batch size 128 for 150 epochs on CIFAR-10, batch size 64 for 100 epochs on Tiny ImageNet. We use cosine scheduler to adjust the learning rate. All the experiments are conducted on Pytorch \cite{torch} framework.

\subsubsection{Defense Settings.} We compare our approaches with the commonly used model repairing methods, \textit{i.e.}, FT, FP \cite{liu2018fine}, NAD \cite{li2020neural} and the SOTA neuron pruning strategy ANP \cite{wu2021adversarial}. All defense methods are allowed to access $1\%$ of the benign training data. Note that \textbf{no data} are used in CLP. The fine-tuning based methods default the training process with batch size 128 and learning rate 0.005 for 20 epochs. We adjust the hyperparameters including pruning ratio in fine-pruning \cite{liu2018fine}, attentional weight in NAD \cite{li2020neural}, and pruning threshold in ANP \cite{wu2021adversarial} to obtain their best performance instructed by their original papers. The CLP hyperparameter $u$ is default to be 3 on CIFAR-10 and 5 on Tiny ImageNet. Further study on the effect of the hyperparameter is conducted in \cref{sec:abl}.

\subsubsection{Evaluation Metric.} The evaluation of the model includes the performance on benign data, Accuracy on Clean data (ACC) and the performance on the backdoored data, which we call the attack success rate (ASR). Note that the ASR is the ratio for poisoned samples that are \textbf{misclassified} as the target label, and it is calculated using the backdoored samples whose ground-truth labels do not belong to the target attack class. In a nutshell, a successful defense should achieve low ASR without much degradation on the ACC. 

\subsection{Experimental Results}

\begin{table*}[htb]
    \centering
    \tiny
    \caption{Performance evaluation of the proposed CLP without data and 4 other defense methods with 500 benign data against seven mainstream attacks on CIFAR-10 with ResNet-18. Higher ACC and Lower ASR are preferable, and the best results are boldfaced. $\downarrow$ denotes the drop rate on average.}
    \begin{tabular}{c|c|cc|cc|cc|cc|cc|cc}
        \hline
        Trigger & \multirow{2}{*}{Attacks} & \multicolumn{2}{c|}{Backdoored} & \multicolumn{2}{c|}{FT} & \multicolumn{2}{c|}{FP\cite{liu2018fine}} & \multicolumn{2}{c|}{NAD\cite{li2020neural}} & \multicolumn{2}{c|}{ANP\cite{wu2021adversarial}} & \multicolumn{2}{c}{CLP(ours)} \cr
         
        Type & & ACC & ASR & ACC & ASR & ACC & ASR & ACC & ASR & ACC & ASR & ACC & ASR \cr
        \hline
        \hline
        \multirow{5}{*}{Static} 
         
        & BadA2O \cite{gu2019badnets} & 93.86 & 100.00 & 92.22 & 2.16 & 92.18 & 2.97 & 91.67 & 5.40 & 91.67 & 5.40 & \bf{93.46} & \bf{1.38} \cr
        
        & BadA2A & 94.60 & 93.89 & 92.03 & 60.76 & 91.75 & 66.82 & 92.86 & 1.33 & 90.29 & 86.22 & \bf{93.69} & \bf{1.06} \cr
        
        & Trojan \cite{Trojan} & 94.06 & 100.00 & 92.58 & 99.99 & 90.78 & 86.43 & 92.13 & 5.76 & 93.44 & 8.11 & \bf{93.54} & \bf{2.06} \cr
        
        & CLA \cite{turner2019label} & 93.14 & 100.00 & 91.86 & 0.39 & 91.02 & 93.21 & \bf{92.46} & \bf{0.44} & 91.13 & 11.76 & 91.89 & 2.84 \cr
        
        & Blended \cite{chen2017targeted} & 94.17 & 99.62 & 93.90 & 70.27 & 90.92 & 3.24 & 92.72 & \bf{1.61} & 93.66 & 5.03 & \bf{94.07} & 1.90 \cr
        
        \hline
        \multirow{2}{*}{Dynamic}
        
        & IAB \cite{nguyen2020input} & 93.87 & 97.91 & 91.78 & \bf{9.52} & 87.04 & 21.33 & 93.52 & 10.61 & \bf{93.52} & 10.61 & 92.78 & 9.88 \cr
        
        & WaNet \cite{nguyen2021wanet} & 94.50 & 99.67 & 92.93 & 9.37 & 92.07 & 1.03 & 94.12 & 0.51 & \bf{94.12} & \bf{0.51} & 94.06 & 0.56 \cr
        
        \hline
        \hline
        
        & Average & 94.03 & 98.72 & 92.47 & 36.07 & 90.82 & 39.29 & 92.78 & 4.30 & 92.54 & 18.23 & \bf{93.36} & \bf{2.81} \cr
        & Drop & $\downarrow$ 0.00 & $\downarrow$ 0.00 & $\downarrow$ 1.56 & $\downarrow$ 62.66 & $\downarrow$ 3.21 & $\downarrow$ 59.43 & $\downarrow$ 1.25 & $\downarrow$ 94.42 & $\downarrow$ 1.49 & $\downarrow$ 80.49 & $\downarrow$ \bf{0.67} & $\downarrow$ \bf{95.91} \cr
        \hline
    \end{tabular}
    \label{tab:cifar10}
\end{table*}

\begin{table*}[htb]
    \centering
    \tiny
    \caption{Performance evaluation of the proposed CLP without data and 4 other defense methods with 1,000 benign data against seven mainstream attacks on Tiny ImageNet with ResNet-18. Higher ACC and Lower ASR are preferable, and the best results are boldfaced. $\downarrow$ denotes the drop rate on average.}
    \begin{tabular}{c|c|cc|cc|cc|cc|cc|cc}
        \hline
        Trigger & \multirow{2}{*}{Attacks} & \multicolumn{2}{c|}{Backdoored} & \multicolumn{2}{c|}{FT} & \multicolumn{2}{c|}{FP\cite{liu2018fine}} & \multicolumn{2}{c|}{NAD\cite{li2020neural}} & \multicolumn{2}{c|}{ANP\cite{wu2021adversarial}} & \multicolumn{2}{c}{CLP(ours)} \cr
         
        Type & & ACC & ASR & ACC & ASR & ACC & ASR & ACC & ASR & ACC & ASR & ACC & ASR \cr
        \hline
        \hline
        \multirow{5}{*}{Static} 
        
        & BadA2O \cite{gu2019badnets} & 62.99 & 99.89 & 56.97 & 99.26 & 57.43 & 57.42 & 61.63 & 0.85 & \bf{63.05} & 3.93 & 62.94 & \bf{0.61} \cr
        
        & Trojan \cite{Trojan} & 64.09 & 99.99 & 62.85 & 3.45 & 60.43 & 99.59 & 61.12 & 95.05 & 62.68 & 10.43 & \bf{63.86} & \bf{0.77} \cr
        
        & CLA \cite{turner2019label} & 64.94 & 84.74 & 53.59 & 27.32 & 61.18 & 82.72 & 59.75 & 30.65 & 60.98 & 15.69 & \bf{64.71} & \bf{0.41} \cr
        
        & Blended \cite{chen2017targeted} & 63.30 & 99.70 & 59.98 & 1.02 & 59.84 & 62.17 & 61.94 & 11.55 & 62.49 & \bf{0.61} & \bf{63.12} & 0.74 \cr
        \hline
        \multirow{3}{*}{Dynamic}
        
        & IAB \cite{nguyen2020input} & \bf{61.40} & 98.28 & 58.35 & 89.89 & 57.03 & \bf{0.21} & 58.17 & 68.43 & 61.39 & 4.67 & 59.09 & 8.70 \cr
        
        & WaNet \cite{nguyen2021wanet} & 60.76 & 99.92 & 57.96 & 97.45 & 53.86 & 23.70 & 56.42 & 87.79 & 54.82 & 86.98 & \bf{59.52} & \bf{1.57} \cr
        
        & SSBA \cite{li2021invisible} & 66.51 & 99.78 & 62.66 & 73.11 & 62.89 & 4.68 & 60.13 & 24.68 & 60.98 & 1.01 & \bf{63.49} & \bf{0.42} \cr
        
        \hline
        \hline
        
        & Average & 63.43 & 97.47 & 58.91 & 55.93 & 58.95 & 47.21 & 59.88 & 62.13 & 60.91 & 17.62 & \bf{62.39} & \bf{1.89} \cr
        & Drop & $\downarrow$ 0.00 & $\downarrow$ 0.00 & $\downarrow$ 4.52 & $\downarrow$ 41.54 & $\downarrow$ 4.48 & $\downarrow$ 50.26 & $\downarrow$ 6.76 & $\downarrow$ 45.57 & $\downarrow$ 2.52 & $\downarrow$ 79.85 & $\downarrow$ \bf{1.04} & $\downarrow$ \bf{95.58} \cr
        \hline
    \end{tabular}
    \label{tab:tinyimagenet}
\end{table*}
In this section, we verify the effectiveness of CLP and compare its performance with other 4 existing model repairing methods as shown in Table \ref{tab:cifar10} and Table \ref{tab:tinyimagenet}. Table \ref{tab:cifar10} shows the experimental results on CIFAR-10, and the proposed CLP remarkably achieves almost the highest robustness against several advanced backdoor attacks. To be specific, the proposed CLP successfully cut the average ASR down to $2.81\%$ with only a slight drop on the ACC ($0.67\%$ on average). Note that CLP reaches such incredible result with no data requirement, and the SOTA defenses ANP and NAD give a similar performance on the ASR with a larger trade-off on the ACC under the access to benign data. 

The standard fine-tuning provides promising defense results against several attacks, especially BadNets (A2O), but fails to generalize to more complex attacks such as Trojan, BadNets (A2A) and blended attack. NAD repairs the backdoored model based on knowledge distillation with supporting information from the attention map of a fine-tuned model. Though it achieves relatively good defense performance, it requires carefully tuning of the hyperparameters.

As for the other pruning based methods, fine-pruning adds an extra neuron pruning step according to the neuron activation to benign images before fine-tuning the model, and achieves the best defense performance against CLA. However, it also fails to maintain high robustness against some more covert attacks. 
ANP utilizes an adversarial strategy to find and prune the neurons that are sensitive to perturbations, which are considered to be highly backdoor related. While both ANP and CLP leverage the concept of sensitivity on channels, ANP measures the sensitivity of the model output to the perturbation on the channels as the pruning index, which requires additional data and careful hyperparameter tuning for different attacks. Unlike ANP, our CLP prunes channels based on the sensitivity of the channels to the inputs, which comes closer to the essence of backdoor attack and can be directly induced by the properties of the model weights without any data. We find both strategy works well on CIFAR-10 against various attacks, but on average, CLP performs better.

\cref{tab:tinyimagenet} shows more experimental results on Tiny ImageNet. All the compared defense methods suffer from severe degradation on both the ACC and the ASR when confront a larger-scale dataset. On the contrast, our CLP still maintains its robustness against those different attacks, including SOTA sample specific attacks IAB and SSBA, which further suggests the strong correlation between channel Lipschitzness and the backdoor behaviors. 

\begin{figure*}[htb]
    \centering
    \includegraphics[width=\linewidth]{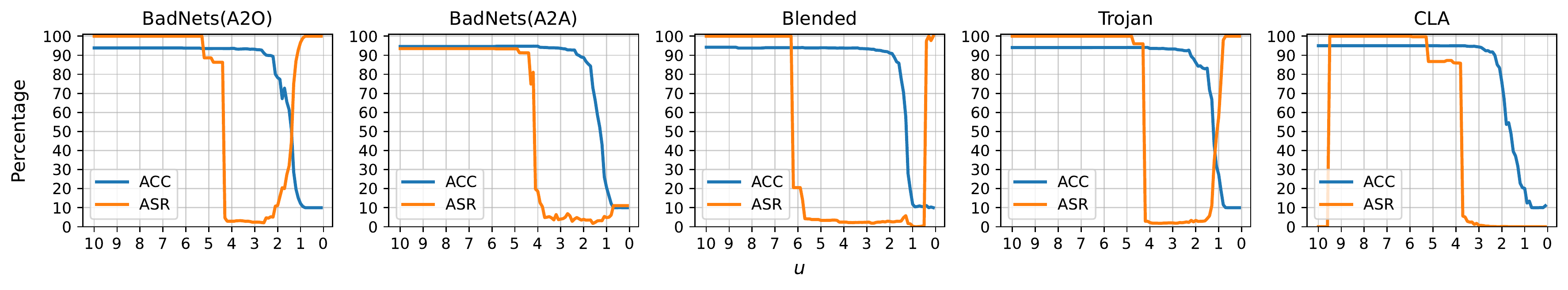}
    \caption{Performance of the CLP with different hyperparameter against different attacks on CIFAR-10 with ResNet-18.}
    \label{fig:attacks}
\end{figure*}
\subsubsection{Performance on Different Architectures.}

\subsection{Ablation Studies}
\label{sec:abl}
\subsubsection{Performance with Different Choices of Hyperparameter $u$.}
As mentioned in \cref{sec:CLP}, the CLP hyperparameter $u$ controls the trade-off between test accuracy on clean data and robust against backdoor attack.
\cref{fig:attacks} shows the ACC and the ASR of the backdoored model after applying CLP with different hyperparameter $u$. 
From the fact that ASR drops rapidly to near $0\%$ while ACC drops much later as $u$ decreases, we argue that the backdoor related channels usually possess higher UCLC than normal channels, which will be pruned precisely by the CLP.
Generally speaking, we can regard the interval between the two points when ASR drops to a very low level and ACC starts to drop as an index of the robustness with hyperparameter. 
For example, it is much easier to choose the hyperparameter $u$ to defend against Blended attack \cite{chen2017targeted} because choosing $u\in[3,5]$ will not affect that much on the performance. 
CLA has the smallest gap, and it requires a more carefully chosen hyperparameter. A possible reason is that the UCLC in the CLA attacked model is not that high compared with other attacked models. Nevertheless, setting $u=3$ still has an acceptable performance on CLA. Note that when $u$ decreases near to 0, nearly all the channels in the model are pruned, so the prediction of the model can be illogical. That's why the ASR curves in some attacks rapidly increase to $100\%$ when $u$ decreases to 0.

\begin{figure}[htb]
    \centering
    \includegraphics[width=0.9\linewidth]{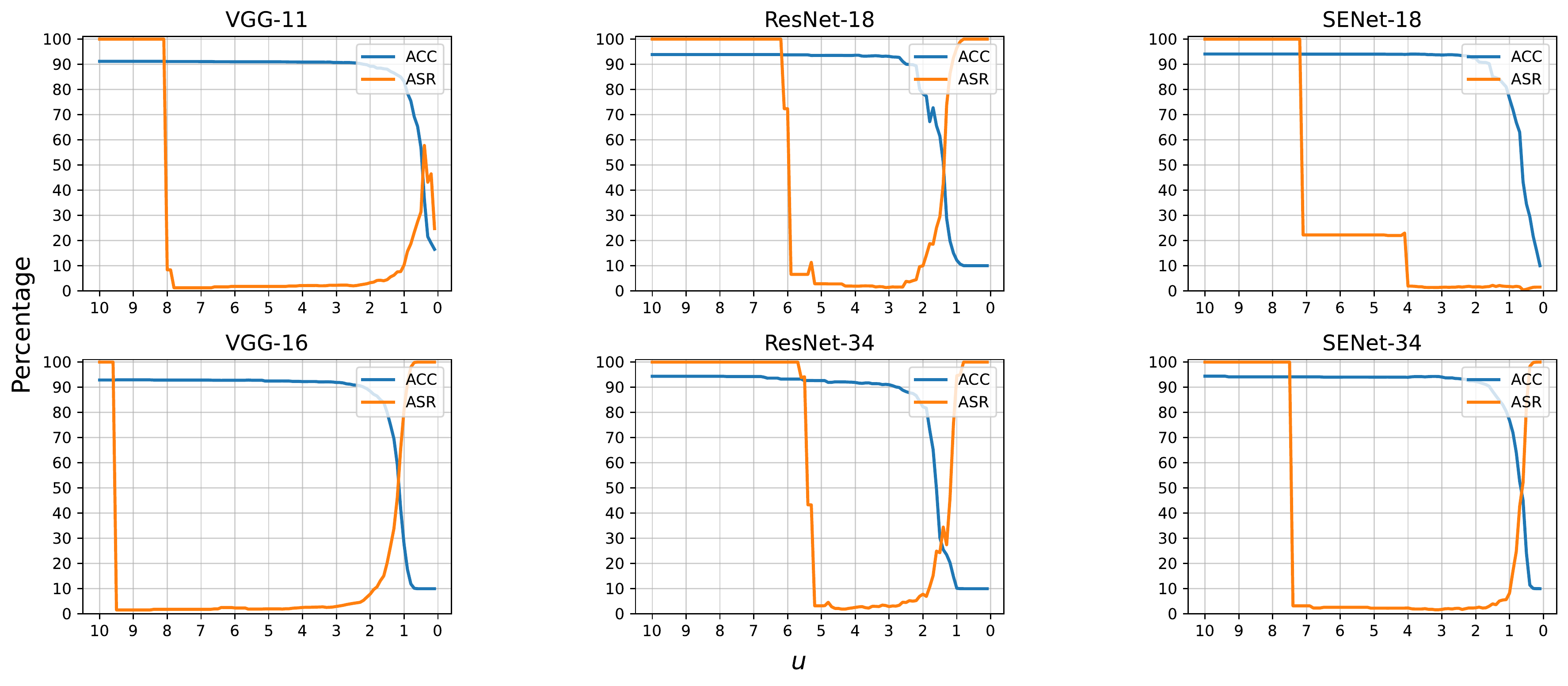}
    \caption{Performance of the CLP with different hyperparameter $u$ in variant CNN architectures against BadNets on CIFAR-10.}
    \label{fig:architectures}
\end{figure}

To verify the generalization performance of CLP across different architectures, we perform BadNets attack on CIFAR-10 with the commonly used CNN architectures ResNet \cite{he2016deep}, VGG \cite{simonyan2014very} and SENet \cite{hu2018squeeze} of different depths. Then we plot the ACC and the ASR curves \textit{w.r.t} to the hyperparameter $u$. This is shown in Figure \ref{fig:architectures}. Overall, CLP achieves very good results on all the tested CNN architectures. Nevertheless, the optimal $u$ for different architectures are different. For example, the optimal choice of $u$ in VGG-16 is about 9. However, such choice of $u$ will not work on other architectures. In addition, we find that both VGG architectures and SENet architectures show better robustness to the choice of hyperparameter than ResNet architectures. In general, choosing $u$ between 3 and 4 generalizes well on different architectures. 

\begin{table*}[htb]
    \centering
    \tiny
    \caption{Performance of the CLP against typical backdoor attacks with different poisoning rate on CIFAR-10 with ResNet-18.}
    \begin{tabular}{c|c|cc|cc|cc|cc|cc}
        \hline
        Poisoning & \multirow{2}{*}{Model} & \multicolumn{2}{c|}{BadNets (A2O)} & \multicolumn{2}{c|}{BadNets (A2A)} & \multicolumn{2}{c|}{Trojan \cite{Trojan}} & \multicolumn{2}{c|}{CLA\cite{turner2019label}} & \multicolumn{2}{c}{Blended\cite{chen2017targeted}}\cr
         
        rate &  & ACC & ASR & ACC & ASR & ACC & ASR & ACC & ASR & ACC & ASR \cr
        \hline
        \hline
        \multirow{2}{*}{1\% } 
        & Backdoored & 93.86 & 100.00 & 94.60 & 93.47 & 94.06 & 100.00 & 93.14 & 100.00 & 94.17 & 100.00 \cr
        & CLP Pruned & 93.46 & 1.38 & 93.69 & 1.06 & 93.54 & 0.92 & 91.89 & 2.84 & 94.07 & 1.90 \cr
        \hline
        
        \multirow{2}{*}{5\% } 
        & Backdoored & 94.29 & 100.00 & 94.21 & 92.57 & 94.48 & 100.00 & 94.46 & 99.86 & 94.33 & 100.00 \cr
        & CLP Pruned & 92.33 & 6.42 & 93.93 & 0.76 & 93.84 & 5.56 & 90.71 & 5.96 & 93.37 & 2.34 \cr
        \hline
        
        \multirow{2}{*}{10\% } 
        & Backdoored & 95.03 & 100.00 & 94.75 & 90.77 & 94.79 & 100.00 & 94.99 & 98.83 & 94.60 & 83.63 \cr
        & CLP Pruned & 94.21 & 2.76 & 94.03 & 0.74 & 93.17 & 3.32 & 93.67 & 10.04 & 93.23 & 0.83 \cr
        \hline
    \end{tabular}
    \label{tab:poison_rate}
\end{table*}

\subsubsection{Performance under Different Poisoning Rates.}
The different choice of poisoning rate also affects the defense performance. To study the robustness of the proposed CLP, we try different poisoning rate of different backdoor attacks with the hyperparameter unchanged ($u=3$ on CIFAR-10 by default). As shown in Table \ref{tab:poison_rate}, CLP effectively reduces the ASR and maintains high ACC under different settings. We note that decreasing the poisoning rate of CLA leads to increasing ASR after applying CLP. CLA with poisoning rate set to $1\%$ gives the worse defense results, and the ASR is about 10\%. However, such ASR doesn't give a considerable threat to our models. Overall, we find poisoning rate doesn't affect much on the performance of CLP.

\subsubsection{Running Time Comparison.}

We record the running time of the above-mentioned defense methods on 500 CIFAR-10 images with ResNet-18, and show the results in \cref{tab:runtime}. The proposed CLP is the only one, which do not require data propagation, so CPU is enough to apply CLP, and we only record the CPU time (i7-5930K CPU @ 3.50GHz). The other methods are evaluated on RTX 2080Ti GPU. The proposed CLP is almost five times faster than the fastest methods among them and only requires 2.87 seconds. 

\begin{table*}[htb]
    \scriptsize
    \centering
    \caption{The overall running time of different defense methods on 500 CIFAR-10 images with ResNet-18. All the methods except for CLP are training on GPU. * denotes that the results of CLP is in CPU time. }
    \begin{tabular}{c|c|c|c|c|c}
    \hline
         Defense Methods & FT & FP & NAD & ANP & CLP* \\
    \hline
    \hline
         Runing Time (sec.) & $\quad$12.35s$\quad$ & $\quad$14.59s$\quad$ & $\quad$25.68s$\quad$ & $\quad$22.08s$\quad$ & $\quad$\textbf{2.87s}$\quad$ \\
    \hline
    \end{tabular}
    \label{tab:runtime}
\end{table*}

\section{Conclusions}
In this paper, we reveal the connection between the Lipschitzness and the backdoor behaviors of the channels in an infected DNN. On this basis, we calculate an upper bound of the channel Lipschitz constant (UCLC) for each channel, and prune the channels with abnormally high UCLC to recover the model, which we refer to as the Channel Lipschitzness based Pruning (CLP).
Due to the fact that UCLC can be directly induced by the model weights, our CLP does not require any data and runs super fast.
To the best of our knowledge, CLP is the first productive data-free backdoor defense method. Extensive experiments show that the proposed CLP can efficiently and effectively remove the injected backdoor while maintaining high ACC against various SOTA attacks.
Finally, ablation studies show that CLP is robust to the \textbf{only} hyperparameter $u$, and generalizes well to different CNN architectures. Our work further shows the effectiveness of channel pruning in defense of backdoor attacks. More importantly, it sheds light on the relationship between the sensitive nature of backdoor and the channel Lipschitz constant, which may help to explain the backdoor behaviors in neural networks.

\section{Acknowledgement}
This work was supported in part by the National Natural Science Foundation of China (No. 62101351), the GuangDong Basic and Applied Basic Research Foundation (No.2020A1515110376), and the Shenzhen Outstanding Scientific and Technological Innovation Talents PhD Startup Project (No. RCBS20210609104447108).

\nocite{langley00}

\bibliography{example_paper}
\bibliographystyle{icml2022}

\newpage
\appendix
\onecolumn
\section{You \emph{can} have an appendix here.}

You can have as much text here as you want. The main body must be at most $8$ pages long.
For the final version, one more page can be added.
If you want, you can use an appendix like this one, even using the one-column format.

\end{document}